\documentclass{article}
\usepackage{spconf,amsmath,epsfig}
\usepackage{booktabs}

\title{Multi-view Remote Sensing Image segmentation with SAM priors}
\name{Zipeng Qi$^{1}$, Chenyang Liu$^{1}$, Zili Liu$^{1}$, Hao Chen$^{3}$, Yongchang Wu$^{2}$, Zhengxia Zou$^{2}$, Zhenwei Shi$^{1,*}$\thanks{Thanks to the National Natural Science Foundation of China under Grant 62125102, the Beijing
Natural Science Foundation under Grant JL23005, the Fundamental Research Funds for the Central Universities, and the National Key Research and Development Program of China (Grant No. 2022ZD0160401). *Corresponding author: Zhenwei Shi (e-mail: shizhenwei@buaa.edu.cn).}}

\address{$^{1}$Image Processing Centre, School of Astronautics, Beihang University, Beijing 100191, China \and $^{2}$Department of Guidance, Navigation and Control, School of Astronautics, Beihang University,
\\Beijing 100191, China \and $^{3}$Shanghai AI Lab}

\begin{document}
%
\maketitle
\begin{abstract}
Multi-view segmentation in Remote Sensing (RS) seeks to segment images from diverse perspectives within a scene. Recent methods leverage 3D information extracted from an Implicit Neural Field (INF), bolstering result consistency across multiple views while using limited accounts of labels (even within 3-5 labels) to streamline labor. Nonetheless, achieving superior performance within the constraints of limited-view labels remains challenging due to inadequate scene-wide supervision and insufficient semantic features within the INF. To address these. we propose to inject the prior of the visual foundation model-Segment Anything(SAM), to the INF to obtain better results under the limited number of training data. Specifically, we contrast SAM features between testing and training views to derive pseudo labels for each testing view, augmenting scene-wide labeling information. Subsequently, we introduce SAM features via a transformer into the INF of the scene, supplementing the semantic information. The experimental results demonstrate that our method outperforms the mainstream method, confirming the efficacy of SAM as a supplement to the INF for this task.
\end{abstract}
\section{Introduction}
\label{sec:intro}
Recent advancements in Remote Sensing (RS) 3D technology, encompassing scene reconstruction~\cite{qi20223d,buyukdemircioglu2022deep}, novel view synthesis~\cite{wu2022remote}, and more ~\cite{gao2022nerf}, ~\cite{qi2022remote}, ~\cite{10283451}, have progressed rapidly. This paper concentrates on the segmentation of images captured from various views within a scene using very small annotations, referred to as multi-view segmentation. This facet holds significance in comprehensively understanding target RS scenes.

Mainstream CNN-based segmentation methods~\cite{cao2022swin, ronneberger2015u, chen2017rethinking, zhou2023occlusion} often rely on extensive labeled training data, which might not be suitable for segmenting multi-view images within an RS scene characterised by a limited number of captured views. Within the development of Implicit Neural Field (INF), ~\cite{qi2023implicit} utilise the INF to encapsulate the colour and density character of each special point in an RS scene. Later, the colour attribute is transformed into a semantic attribute through a limited number of supervisions (e.g., 3–5 semantic labels). When the above process is optimised, we sample points along rays passing through the camera's centre and each pixel in the input image, and compute the colour and semantic class (seg.) of each pixel using the colour rendering and semantic rendering functions. The density, colour, and semantic attributes of the INF encode the scene's 3D, colour, and semantic information, respectively. 


Nevertheless, the semantic attributes of certain sampled points might be under-fitted due to the limited coverage of the entire scene by a restricted number of semantic annotations. 
\begin{figure*}[ht]
\centering
\includegraphics[width=\linewidth]{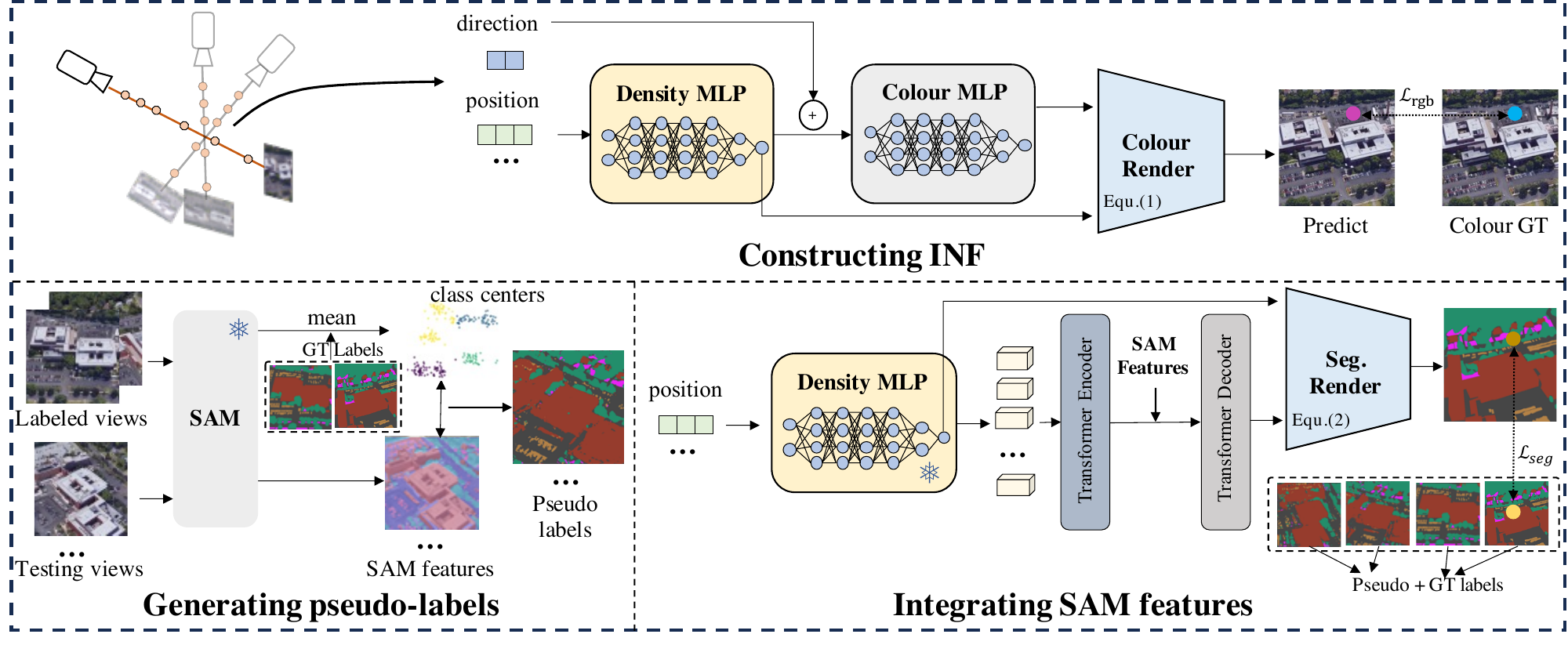}
\caption{Our proposed method consists of two stages. Initially, we employ two MLPs to build the scene's INF, encoding 3D information in the density attributes of each spatial point, supervised by all RGB images. Subsequently, we freeze the density MLP and incorporate SAM priors into the INF. This involves transferring the colour attribute by introducing pseudo-labels as additional supervision and injecting SAM features via a transformer. }
\label{method}
\end{figure*}
In this paper, we propose to integrate the prior knowledge of a large foundation model into the INF to tackle the aforementioned challenges. We specifically selected a visual foundation model-Segment Anything(SAM)~\cite{kirillov2023segment} due to its extensive training on diverse data, enabling it to segment arbitrary input images. Its encoder demonstrates the capacity to robustly extract image features, even for RS images. Our proposed method comprises two stages: (1) constructing the scene's INF, and (2) transferring the colour attribute into the semantic attribute. We introduce SAM priors into the INF using a transformer mechanism during the second stage to augment semantic information. Additionally, we compare SAM features between testing and training views to derive pseudo-labels for each testing view, promoting scene-wide supervision. 

In Section~\ref{sec:method}, we give a detailed introduction to the proposed method. In Section~\ref{sec:experiment}, the experimental results are presented. Conclusions are drawn in section~\ref{sec:concution}.

\section{METHODOLOGY}
\label{sec:method}
\subsection{Overview}
As depicted in Fig.~\ref{method}, our proposed method consists of two stages. In the first stage, we extract the 3D information of the RS scene encoded in the density attribute of the INF, utilising all RGB images. The second stage involves transferring the colour attribute of each sampled point to the semantic attribute using a limited number of training labels. Additionally, we leverage SAM features to generate pseudo-labels for the testing views, thereby enhancing the semantic supervision of the scene. Furthermore, we introduce SAM features into the INF through a transformer mechanism to enrich the semantic information.

\subsection{Constructing INF}
In the first stage, 
we utilise the pixel colour from all input RGB images to supervise the construction of the scene's INF. Specifically, we randomly select rays ($R$) originating from the camera centre and passing through pixels in the image plane, sampling points along these rays. Two MLPs are employed, taking the point position and ray direction as input to predict the colour and density attributes for each sampled point. Subsequently, we employ the following function to render the colour of each pixel. 
\begin{equation}
\label{equ:1}
    \text{colour} = \sum \limits ^{N}_{i=1} \exp{\Big(-\sum \limits ^{i-1}_{j=1}\alpha_j\sigma_j\Big)}\Big(1-\exp{(-\alpha_i\sigma_i})\Big)c_i,
\end{equation}
where $\sigma_i$ is the density attribute of the $i_{th}$ sampling point, $c_i$ is the colour attribute and $\alpha_j$ is the interval distance between the $i$ point and the
$i+1$ point. Then, we use the following loss to optimise the parameters of the two MLPs:
\begin{equation}
\label{equ:3}
    \text{$loss_{rgb}$} = \sum \limits_{r \in R}\Big[\Vert \text{$colour_r$}- \text{$GT_r$}\Vert_2^2\Big].
\end{equation}

\subsection{Transfering semantic attribute}
In the second stage, our proposal involves integrating SAM priors into the process of converting the colour attribute into the semantic attribute of each sampled point.

\textbf{Generating pseudo labels.} While the INF mechanism effectively utilises semantic information from a limited training dataset, optimising semantic attributes for points beyond the coverage of the training view remains a challenge. To address this, we utilise SAM features to generate pseudo-labels for testing views. Initially, we compute SAM encoder features for each view's images using the equation:
\begin{equation}
\label{equ:4}
    \text{$F_i$} = \text{SAM}(\text{$image_i$}).
\end{equation}
Subsequently, we calculate the mean value of the features (F) associated with pixels from the training data belonging to category $c$—serving as the center of category $c$ ($c = 1 \dots L$, where L represents the number of classes). Finally, we compare features corresponding to each pixel of testing images with the computed centers to assign a category based on the minimum distance. The results we take as pseudo-labels, as shown in the left-down of Fig.~\ref{method}.

\textbf{Intergrating SAM features via Transformer.} Directly converting the colour attributes into semantic attributes lacks texture information, resulting in poor results. Thus, we use the $F_i$, which contains rich and very robust semantic information, as the local texture prompter.
\begin{equation}
\begin{aligned}
s_1, \dots, s_n &= T_e(b_1, \dots, b_n), \\
s_1^1, \dots, s_n^1 &= T_d(\{s_1, \dots, s_n\}, F_i^{x,y}),\\
s_1^2, \dots, s_n^2 &= s_1 + s_1^1, \dots s_n + s_n^1,\\
\label{equ:5b}
\end{aligned}
\end{equation}
Here, the $b_i$ is the base feature of the $i_{th}$ sampling point along the ray $r$ for generating the density and colour attributes, $s_i^2$ represents the semantic attribute, and ${x,y}$ denotes the coordinates of the intersection point of the ray $r$ with the image. $T_e$ refers to the transformer encoder, and $T_d$ signifies the transformer decoder. In the aforementioned process, the encoder is responsible for transforming the base features of sampling points into semantic features, while the decoder is designed to integrate SAM features with the semantic features. We generate the semantic label using the following function.
\begin{equation}
\label{equ:2}
    \text{seg} = \sum \limits ^{N}_{i=1} \exp{\Big(-\sum \limits ^{i-1}_{j=1}\alpha_j\sigma_j\Big)}\Big(1-\exp{(-\alpha_i\sigma_i})\Big)s_i^2,
\end{equation}
where $s_i^2$ is the semantic attribute of the $i_{th}$ sampling point. Then we employ the subsequent loss function to compute the loss:
\begin{equation}
\label{euq:6}
    \mathcal{L}_{s} = -\lambda\sum \limits_{r \in R} \Big[ \sum \limits_{l=1}^{L} seg_r^l\log{GT_r^l}\Big],
\end{equation}
$L$ is the number of classes. When $R$ belongs to the training views, the value of $\lambda$ is set to 1. When $R$ belongs to the testing views, we utilise the pseudo-labels as the ground truth and set $\lambda$ to 0.001.

\section{Experiment}
\label{sec:experiment}

\subsection{Experimental Setup}
We perform experiments on the real sub-datasets introduced in~\cite{qi2023implicit}. The whole real sub-dataset includes 300 images, each sized at $512 \times 512$ pixels. Merely 2\% of the images in the training sets possess corresponding labels. The INF model in our method is based on the NeRF++~\cite{zhang2020nerf++} and our model is implemented on PyTorch and trained using a single NVIDIA RTX 3090 GPU. The comparison methods we selected comprise two categories: CNN-based and INF-based methods, including Unet~\cite{ronneberger2015u}, SegNet~\cite{badrinarayanan2017segnet}, DANet~\cite{fu2019dual}, DeepLab~\cite{chen2017rethinking}, SETR~\cite{zheng2021rethinking} and Sem-NeRF~\cite{zhi2021place}. All these methods were re-trained using the previously mentioned dataset. We use average mIOU across all sub-datasets as the validation metric.

\subsection{ Experiment Results}
\begin{figure}
    \centering
    \includegraphics[width=\linewidth]{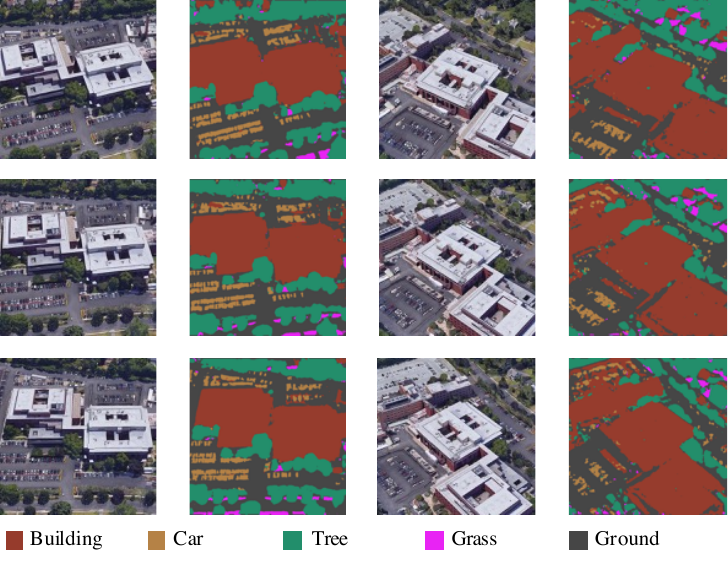}
    \caption{The images on the left showcase results in proximity to the training views, while those on the right depict regions distant from the training views.}
    \label{fig:results}
\end{figure}

\begin{table}[!ht] 
\centering
\caption{MIoU metric of different methods on sub-datasets.}
\label{metric-results}
\resizebox{\linewidth}{!}{
\begin{tabular}{c|cccccccc}
\toprule
Methods  & SegNet & Unet & DANet & Deeplab & SETR & Sem-NeRF  & Ours  \\\midrule
real \#1 &8.67& 64.94  & 35.71 & 39.35 & 36.26 & 11.64  & 62.29 \\ 
real \#2 & 27.88  & 49.95 &49.13 & 41.37 & 31.04 & 18.78 & 79.94 \\ 
real \#3  & 18.84 & 68.33  & 34.30  &  52.37 & 26.07 & 19.76 & 61.93  \\ \midrule
AVG & 18.46 & 61.07 & 39.71 & 44.36 & 31.12 & 16.72 & \textbf{67.95} \\\bottomrule
\end{tabular}
}
\end{table}

\textbf{Quantitative Results:}
We present the quantitative comparison results between our method and other comparison methods in Table~\ref{metric-results}. Compared with CNN-based and INF-based methods, our segmentation accuracy has been improved to a certain extent in the avg. mIoU metric. Compared with CNN-based methods, due to the lack of spatial information, the consistency between views of the results of these methods is very poor. Compared to Unet, our method demonstrates an improvement of over 6.8\% on the avg.mIOU metric. Additionally, when compared to Sem-NeRF, which is not suitable for modeling remote sensing scenes due to significant differences between distant and close views, our method outperforms it by over 50\%. The experimental results show that our proposed method is more friendly to large-scale remote sensing scenes, and we introduce SAM features into multi-view segmentation tasks for the first time, verifying its effectiveness.

\textbf{Quantitative Results:} Fig.~\ref{fig:results} illustrates the multi-view segmentation results of the real1 sub-dataset. The proposed method accurately segments images from the test view using annotations from only two training views, ensuring strong inter-view consistency. Notably, our method exhibits commendable segmentation performance in regions far from the training perspective on the right part of the figure. This is achieved by leveraging pseudo-labels and the feature conspicuousness of SAM, demonstrating the effectiveness of our approach.

\section{Conclusion}
\label{sec:concution}
This paper focuses on multi-view segmentation within RS scenes, aiming to tackle challenges arising from insufficient scene-wide supervision and inadequate semantic features within the implicit neural field. To address these, we leverage SAM to derive pseudo-labels for each test view, thereby enhancing scene-wide semantic supervision. Additionally, we integrate SAM features into the INF using a transformer to bolster semantic information. Experimental results substantiate that our method outperforms the CNN-based and INF-based method, validating the efficacy of SAM as a valuable supplement to the INF for this task.

\bibliographystyle{IEEEbib}
\bibliography{refs}

\end{document}